\documentclass[sigconf, screen, authorversion]{acmartmodified}

\setcopyright{rightsretained}
\copyrightyear{2024}
\acmYear{2024}
\acmDOI{}

\acmConference[GenAI Evaluation KDD2024]{KDD workshop on Evaluation and Trustworthiness of Generative AI Models}{August 25, 2024}{Barcelona, Spain.}
\acmBooktitle{
}
\acmISBN{}

\newcommand{\D}{\mathcal{D}}
\newcommand{\Dtrain}{\D^\text{train}}
\newcommand{\Dval}{\D^\text{val}}
\newcommand{\Dtest}{\D^\text{test}}
\newcommand{\Dref}{\D^\text{ref}}

\newcommand{\xnew}{p^\text{new}}
\newcommand{\mnew}{m^\text{new}}
\newcommand{\ntrain}{n_\text{train}}

\newcommand{\R}{\mathbb{R}}
\renewcommand{\L}{\mathcal{L}}
\newcommand{\Ltrain}{\L^\text{train}}
\newcommand{\Lval}{\L^\text{val}}
\newcommand{\Ltest}{\L^\text{test}}
\usepackage{graphicx}
\usepackage{subcaption}
\usepackage{array} %
\usepackage{fancyhdr} %
\usepackage{adjustbox}

\begin{document}

\title[100 instances is all you need]{100 instances is all you need: predicting the success of a new LLM on unseen data by testing on a few instances}

\author{Lorenzo Pacchiardi}
\email{lp666@cam.ac.uk}
\affiliation{%
  \institution{Leverhulme Centre for the Future of Intelligence, University of Cambridge}
  \city{Cambridge}
  \country{UK}
}

\author{Lucy Cheke}
\email{lgc23@cam.ac.uk}
\affiliation{%
  \institution{Department of Psychology, Leverhulme Centre for the Future of Intelligence,, University of Cambridge}
  \city{Cambridge}
  \country{UK}
}

\author{José Hernández-Orallo}
\email{jorallo@upv.es}
\affiliation{%
  \institution{Leverhulme Centre for the Future of Intelligence, University of Cambridge}
  \city{Cambridge}
  \country{UK}
}
\affiliation{%
  \institution{VRAIN, ValGRAI, Universitat Politècnica de València}
  \city{València}
  \country{Spain}
}

\begin{abstract}
Predicting the performance of LLMs on individual task instances is essential to ensure their reliability in high-stakes applications. To do so, a possibility is to evaluate the considered LLM on a set of task instances and train an \textit{assessor} to predict its performance based on features of the instances. However, this approach requires evaluating each new LLM on a sufficiently large set of task instances to train an assessor specific to it. %
In this work, we leverage the evaluation results of previously tested LLMs to reduce the number of evaluations required to predict the performance of a new LLM. In practice, we propose to test the new LLM on a small set of reference instances and train a \textit{generic assessor} which predicts the performance of the LLM on an instance based on the performance of the former on the reference set and features of the instance of interest. We conduct empirical studies on HELM-Lite and KindsOfReasoning, a collection of existing reasoning datasets that we introduce, where we evaluate all instruction-fine-tuned OpenAI models until \texttt{gpt4-0125-preview}. 
When predicting performance on instances with the same distribution as those used to train the generic assessor, we find this achieves performance comparable to the LLM-specific assessors trained on the full set of instances. 
Additionally, we find that randomly selecting the reference instances performs as well as some advanced selection methods we tested.
For out of distribution, however, no clear winner emerges and the overall performance is worse, suggesting that the inherent predictability of LLMs is low. 
\end{abstract}

\ccsdesc[500]{Computing methodologies~Natural language processing}
\ccsdesc[300]{Computing methodologies~Transfer learning}
\ccsdesc[100]{Computing methodologies~Cost-sensitive learning}
\ccsdesc[500]{General and reference~Evaluation}
\ccsdesc[500]{General and reference~Reliability}
\keywords{Large language models, evaluation, performance prediction, predictable AI.}
\begin{teaserfigure}
\centering
  \includegraphics[width=0.8\textwidth]{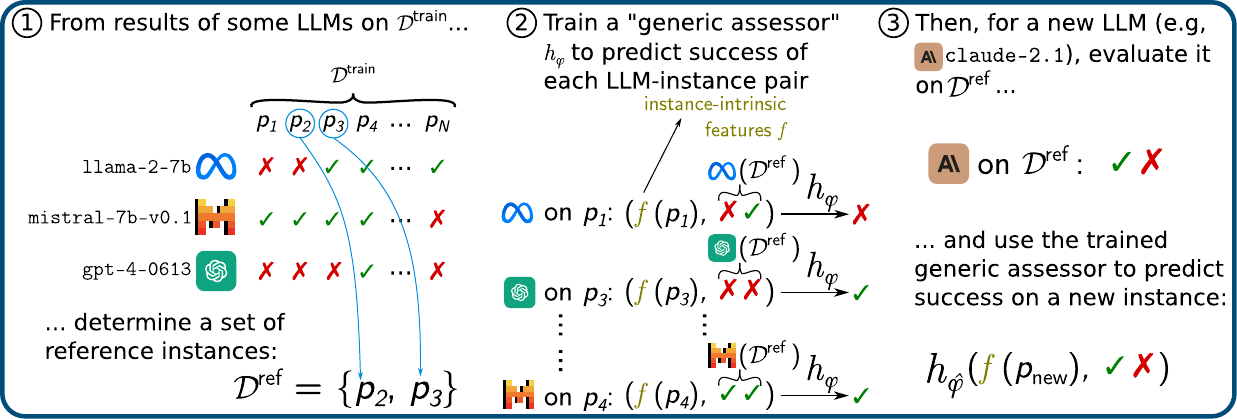}
  \caption{Our proposed pipeline for predicting the performance of a new LLM on a new instance by testing on a few instances: starting from instance-level evaluation results of a set of LLMs, a reference set of instances is extracted (1). Then, we train a ``generic assessor'' that predicts the performance of each LLM-instance pair, based on features intrinsic to the instance (e.g., vector embeddings) and the performance of the considered LLM on the reference set (2). The performance of the new LLM on a new instance can be predicted by evaluating the new LLM on the reference set and applying the trained generic assessor (3).}
  \Description{}
  \label{fig:teaser}
\end{teaserfigure}

\received{20 February 2007}
\received[revised]{12 March 2009}
\received[accepted]{5 June 2009}

\maketitle

\section{Introduction} \label{sec:intro}

Large Language Models (LLMs) are being used as components of multiple services and products, such as agents performing general computer tasks \cite{kim2024language},  performing ML experiments \cite{huang2024benchmarking}, and even operating unmanned aerial vehicles \cite{javaid2024large}. These systems typically query an LLM on a specific instance of a task and use their output to determine a sequence of actions. For some of these uses, it is essential to determine whether the output produced by the LLM on a specific task instance is correct (or, more generally, ``valid'' \cite{zhou2023predictable}) before the subsequent steps are executed\footnote{Notice that this cannot rely on the ``ground truth'' of the task instance, as that is not available in practical use cases (otherwise, there would be no need to query the LLM).}. A nascent line of research \cite{zhou2022reject, hernandez2022training} is addressing this problem 
by developing ``assessors'', namely, independent modules that predict the correctness (or a continuous performance score) of an AI system on an instance based on features intrinsic to the latter (such as linguistic features or sentence vector embeddings). Assessors can be specific to an AI system, or ``generic'', in which case they also take as input information on the AI system at hand and are trained to predict the performance of different LLMs on different instances.

Meanwhile, the rate at which new LLMs are released has drastically increased. Some providers, such as OpenAI, are iteratively retiring old versions when new ones are released, forcing developers to update the LLM version used in their applications (see \cite{openai_deprecations}). An even larger explosion is occurring in the open-source world, fuelled by inexpensive fine-tuning techniques \cite{hu2022lora}. To build an assessor specific to a new LLM version, users must evaluate it on a sufficiently large set of task instances, causing the costs to rise quickly when considering many LLM versions. On the other hand, the system information one might use to build a generic assessor, such as the number of parameters or statistics of the training data or architecture, is not standardised across LLMs and unavailable for proprietary models.

As such, this paper investigates the following question: 
\textbf{can we combine information across LLMs to predict the performance of a new LLM on a new instance by relying only on observational (or behavioural) features of the LLMs?}
In practice, we propose to characterise each LLM by its performance on a small set of \textit{reference instances} and to build a generic assessor using those as system features.
More precisely, we first select a small set of reference instances from the labelled dataset on which past LLMs were evaluated. Then, we train the generic assessor on the concatenation of instance-specific features and the LLM-specific success vector on the reference instances. Finally, to estimate the probability of success of a new LLM on a novel instance, it suffices to evaluate the former on the reference instances, concatenate its performance to the features of the instance, and apply the trained generic assessor.

In our empirical studies, we rely on HELM-Lite \cite{helm_lite}, which provides instance-level results for 30 LLMs from different providers (at the time we conducted our experiments), and a collection of previously existing datasets we introduce, named ``KindsOfReasoning'', on which we evaluated the full set of instruction-following models from OpenAI until \texttt{gpt4-0125-preview}. We only consider tasks with binary correctness score (thus discarding the datasets in HELM-Lite that do not satisfy this) and therefore build binary assessors.

We train specific assessors using different prompt features and find that OpenAI embeddings \cite{openai_embeddings} lead to better in-distribution performance than simpler methods such as \texttt{Word2vec} \cite{Mikolov2013EfficientEO}, \texttt{FastText} \cite{bojanowski2017enriching}, and n-grams. Although this analysis is not the main focus of our work, it is a valuable tangential contribution.
Subsequently, we build generic assessors using various methods to select the reference instances and combine the performance on these with the instance-specific features. 
When predicting performance on instances with the same distribution as those used to train the generic assessor, we find the latter to perform comparably to the specific assessors, which require the LLM to be evaluated on many more instances. 
Additionally, we find that a random selection of reference instances performs as well as the advanced selection methods we tested.
However, in out-of-distribution scenarios, the predictive power of all assessors declines significantly, indicating a lack of general predictability in LLMs.

In essence, the main contributions of our work are the following:
\begin{itemize}
    \item We propose a framework that combines evaluation results across LLMs to predict the performance of a new LLM on a new instance by only evaluating the new LLM on a small set of reference instances.
    \item We study the performance of various methods for selecting the reference instances and combining their performance with instance-specific features to build the generic assessor
    \item Finally, we introduce the KindsOfReasoning collection of existing datasets testing various kinds of reasoning and, in the spirit of making instance level results available \cite{burnell2023rethink}, we publicly release the raw outputs and the evaluation results of all instruction-tuned models from OpenAI. To the best of our knowledge, this is the first public release of its kind.
\end{itemize}

\looseness=-1
The rest of the paper is organised as follows: Section \ref{sec:related} reviews related works in the area of predicting the performance of large language models (LLMs). In Section \ref{sec:methods}, we describe our methodology. %
Section \ref{sec:experiments} presents our empirical studies, where we compare the performance of the generic assessor with that of independent assessors, and study how well the generic assessor can select the most suitable LLM for a task. %
In Section \ref{sec:conclusion}, we conclude the paper and discuss the limitations of our study and directions for future work (\ref{sec:limitations}).

\section{Related work} \label{sec:related}

\subsection{Instance-level prediction of success of AI systems}

The motivation for our work follows \cite{zhou2023predictable}, which advocates for the importance of instance-level success predictions for AI systems and coins the term ``predictable AI''; in particular, they highlight how ensuring predictability should be prioritised over increases in average performance for risky use cases, and how having this could help with compliance with upcoming regulatory frameworks, such as the EU AI Act \cite{benifei2023proposal}. 

\looseness=-1
Following this motivation, \cite{hernandez2022training} introduces the concept of an \textit{assessor model}, which accompanies an ML system and estimates the probability of success of the system on individual instances. In particular, they discuss how an assessor can be trained on the evaluation results of the ML system on test data (i.e., which has not been used for training the ML system). Finally, \cite{zhou2022reject} applies this idea to LLMs, by showing how a smaller external model can be used to predict the performance of a bigger and more expensive LLM on individual instances without passing the latter through the model. They also find it possible to reject almost half of the failure cases before running much larger LLMs, resulting in a significant saving of compute.

\subsection{Predictability of aggregated benchmark scores from LLM-specific features}

Two works \citep{ye2023how,owen2024predictable} studied the extent to which an LLM's aggregate performance on  BIG-Bench tasks \cite{srivastava2022beyond} can be predicted using information on the LLM such as number of parameters or the amount of used compute. In contrast, our work does not rely on these quantities, which are often unavailable, instead characterising LLMs according to their performance on reference samples. Moreover, while these works focus on predicting aggregate performance, our work and the ones mentioned in the previous subsection provide instance-level predictions for new unlabelled instances.

\subsection{Extracting LLM-specific features from existing evaluations}
\looseness=-1
Recently, \cite{ruan2024observational} built ``observational scaling laws'' that link performance on complex downstream tasks to hypothesised latent capabilities, whose values can be inferred by decomposing the performance of various LLMs on different benchmarks into components linked by a log-linear relation with compute measures for LLM training. Doing so allows us to combine information across different LLM families, which may differ for their efficiency in converting raw compute into benchmark scores. Once this relation is established, the performance of a new model on downstream tasks can be predicted by knowing its performance on simple benchmarks and its compute cost. \cite{ruan2024observational}  also select a subset of LLMs that maintains high prediction accuracy while reducing cost by requiring the evaluation of performance on downstream tasks for fewer models.
Their work is similar to ours in determining LLM-specific features by using evaluation results of multiple LLMs and using them to predict the performance of a new LLM. However, the aim of our work is to predict the performance of the new LLM on a specific instance with as few evaluations as possible, while the aim of \cite{ruan2024observational} is to avoid the cost of evaluating complex downstream tasks and predict the performance on the latter from results on simple benchmarks and compute measures. As such, the LLM-specific features they use (the latent capabilities) are obtained from the performance of the new LLM on simple benchmarks (which \cite{ruan2024observational} assumes to be available), while our method only needs to evaluate the LLM on a small number of instances. Moreover, our method can be applied to new instances for which no ground truth is available, while the simple benchmarks and the downstream tasks employed in \cite{ruan2024observational} must have a grading mechanism.

\subsection{Predicting performance by benchmark subsampling }
Several works share the motivation of reducing the number of evaluations (and hence the cost) needed to evaluate a LLM. For instance, a ``Lite'' version with a reduced number of tasks was introduced alongside the larger BIG-Bench benchmark \citep{srivastava2022beyond}; however, the way in which the task selection was done is unknown, to the best of our knowledge. Similarly, HELM-Lite \cite{helm_lite} is a revised and reduced version of HELM \citep{liang2022holistic}.  
However, both of these perform the reduction at the level of \textit{tasks} of which the benchmark is constituted. 
Instead, \cite{vivek2024} subsample a dataset by clustering models' confidence to predict the overall accuracy on the whole dataset, while MixEval \cite{ni2024mixeval} extracts a subset of instances from various benchmarks which is most predictive of the performance on Chatbot Arena\footnote{\url{https://chat.lmsys.org/}}, an online platform performing pairwise comparison of LLM outputs. 
Closer to our work is TinyBenchmarks \citep{polo2024tinybenchmarks}, which selects informative \textit{instances} from HELM-Lite and estimates the performance of a new LLM on the whole benchmark by evaluating it only on those instances. In particular, TinyBenchmarks uses Item Response Theory (IRT) on the matrix of success of each LLM on each instance present in the HELM-Lite dataset to extract a vector of item demands and LLM capabilities. Then, it uses either the item demands or the raw LLM success on each instance to build a representative subset of instances by clustering the items and taking the cluster centroids. Similarly to our approach, a new LLM is then only evaluated on the representative subset; however, in contrast to our work, they aim to predict the aggregate score on the benchmark, while we focus on predicting instance-level performance. In practice, their IRT method provides instance-level predictions (which the authors aggregate), but these predictions are limited to instances on which previous LLMs have been evaluated (as this is necessary to obtain the item demands), which requires access to the ground truth. In contrast, our approach is applicable to new instances for which we do not know the ground truth, as the trained assessor does not require any information beyond the intrinsic features of test instances.
A similar work to \cite{polo2024tinybenchmarks} is \texttt{metabench} \cite{kipnis2024metabench}, which considered 6 different datasets, and performed a two-step procedure (random sampling for each dataset, followed by item selection based on the Fisher information matrices of IRT item parameters) to extract a small set of instances, the performance on which accurately predicts aggregate performance on the 6 datasets. As they fit the IRT model only the pre-selected instances, their method is unable to predict instance-level performance.
Finally, despite not tackling predictability directly, \cite{ailem2024} finds that the vector of successes of different LLMs is correlated across instances belonging to 4 benchmarks, and, for one of those benchmarks, the similarity between the embeddings or a pair of instances predicts the similarity between the success vectors; this suggests that patterns in success across LLMs can be found and related to the embeddings.

\subsection{Evaluations of reasoning in LLMs}

\cite{burnell2023revealing} found reasoning to be one of three factors in the capabilities of LLMs. Indeed, reasoning in LLMs has been extensively studied: see \cite{mondorf2024beyond} for a survey on LLM reasoning evaluations and \cite{huang-chang-2023-towards} for a broader survey also encompassing ways to improve and elicit reasoning in LLMs. 

Recently, several collections of reasoning datasets have been introduced. GLoRE \citep{teng2023glore} collects 12 logical reasoning datasets with three different types of tasks (multiple choice, natural language inference, and binary answers). Similarly, LogiGLUE \citep{luo2023towards} collects 24 datasets related to inductive, deductive and abductive reasoning, with four different types of tasks (the same ones as GLoRe and free-form question answering); they only selected datasets that do not require external domain knowledge, but some of these datasets are poorly formatted. Finally, CALM-Bench \citep{dalal2023calm} is a collection of 6 diverse tasks requiring both causal reasoning and knowledge. KindsOfReasoning, the collection we introduce combining previously existing datasets testing various kinds of reasoning, partly overlaps with each of the aforementioned collections; however, KindsOfReasoning aims to include a broader range reasoning types (logical, common sense, inductive, deductive, abductive, counterfactual, causal, analogical, spatial and arithmetic reasoning) over 22 different datasets; see Appendix~\ref{app:kindsofreasoning} for more information on the dataset construction.

\section{Methodology}
\label{sec:methods}

Let us denote by $\mathcal L =\{m_j, j=1,\ldots, n \}$, a set of trained LLMs. Moreover, let $\D = \{ (p_i, y_i), i=1,\ldots,N\} $ be a test dataset used to evaluate the performance of the LLMs, with $i$ denoting instance index, $p_i$ the input to the LLM (i.e., the prompt) and $y_i$ the target value (i.e., the expected completion by the LLM). Further, we will denote by $m_j(p_i)$ the output $m_j$ produces when given $p_i$ as input\footnote{As LLMs are stochastic, $m_j(p_i)$ is in general a random variable, and so is $z_{j,i}$. In our empirical study, we sample the LLMs at 0 temperature, but, even so, there is still a residual amount of stochasticity, even though the reason for this is unclear \cite{openai_0T}.} and by $z_{j,i}$ a binary value indicating the ``correctness'' of $m_j(p_i)$ with respect to $y_i$. The correctness $z_{j,i}$ can be defined in multiple manners (for instance, exact match or whether $y_i$ is a substring of $m_j(p_i)$); the most suitable manner depends on the considered task, but in general the aim of $z_{j,i}$ is to capture what a human judge would perceive as a correct answer\footnote{Particularly in the case of free-form question answering, it can be tricky to find a formulation that always matches what a human judge would perceive as a correct answer.}.

In the following, we first frame the problem of predicting the correctness $z_{j,i}$ and then discuss our main contribution, 
namely a framework to predict the performance of a new LLM by evaluating it on a small subset of instances. 

\subsection{Predicting success of a LLM using features intrinsic to the prompt}
\label{sec:full_assessor}
To begin with, let us now consider a single LLM, say $m_1$; our aim is to train a classifier (termed ``assessor'') to predict the performance $z_{1,i}$ from the prompt $p_i$.
To do so, we split the test dataset $\D$ into different splits used to train, validate and evaluate the assessor \cite{hernandez2022training}, denoted as $\Dtrain, \Dval$ and $\Dtest$, such that $\D = D^\text{train} \cup \Dval \cup\Dtest $ and $\Dtrain \cap \Dval = \Dval \cap \Dtest = \Dtrain \cap \Dtest = \varnothing$. In a real-world scenario, $\Dtest$ will represent instances for which we did not evaluate the considered LLM (and for which we may not have access to the ground truth); in contrast, available evaluation results are split into $\Dtrain$ and $\Dval$.

In practice, we can extract some numerical features $ f(p_i)$ from the textual prompt $p_i$; we use ``intrinsic'' features, i.e. features that are defined independently of the problem at hand (such as the number of negations or the vector embeddings of the sentence). Formally, we consider a loss function $\ell$ and a family of classifiers $h_\varphi$, where $\varphi$ denotes the parameters of the classifier (for instance, the weights in a logistic regression classifier), and aim to minimise
\begin{equation}
\sum_{p_i\in\Dtrain} \ell(h_\varphi(f(p_i)), z_{1,i})
\end{equation}
over $\varphi$ using some optimisation algorithm; we can then select the best hyper-parameters for solving the above problem using the performance on the validation data $\Dval$, leading to picking a classifier $h_{\hat \varphi}$. Now, we can predict the performance of $m_1$ on $\xnew\in\Dtest$ as $h_{\hat \varphi}(f(\xnew))$ without inputing the prompt $\xnew$ into the LLM $m_1$\footnote{This assessor is anticipative \cite{hernandez2022training}, as it does not use the output $m_1(\xnew)$ when predicting the performance; this can avoid the cost of querying the LLM if its performance on a specific input is predicted to be poor.}.

\subsection{Predicting success by evaluation on reference instances}
\label{sec:reference}
Now, consider the case in which we have previously evaluated some LLMs on $\Dtrain$ and $\Dval$. We are interested in predicting the performance of a new LLM, say $m^\text{new}$ on new instances $\Dtest$. Using the approach in Section~\ref{sec:full_assessor}, we could test the new LLM on all instances in $\Dtrain$ and $\Dval$ and train an assessor specific to that LLM. Instead, we want to leverage the information contained in the available evaluation results for previous LLMs to predict the performance of $m^\text{new}$ on $\Dtest$ without evaluating it on the full $\Dtrain$ (and assuming that we do not have access to the labels in $\Dtest$, which prevents us from evaluating the other LLMs on it). 

Thus, we build a \textit{generic assessor}, namely a classifier that predicts the success $z_{j,i}$ from the pair $(m_j,p_i)$. In practice, let us split the LLMs for which full evaluation results are available into a training and validation split $\Ltrain$ and $\Lval$. 
For each pair $(m_j,p_i) \in \Ltrain \times\Dtrain $, we concatenate the prompt-intrinsic features $f(p_i)$ with LLM-specific features $g(m_j)$ and aim to fit a classifier $h_\varphi$ that minimises
\begin{equation}\label{Eq:joint_assessor}
\sum_{m_j\in\Ltrain} \sum_{p_i\in\Dtrain} \ell(h_\varphi(g(m_j), f(p_i)), z_{j,i})
\end{equation}
over $\varphi$. Similarly to what we did before (Section~\ref{sec:full_assessor}), we keep the performance of $\Lval$ on $\Dval$ to perform model selection, leading to a trained classifier $h_{\hat \varphi}$. Then, the performance of $m^{\text{new}}$ on an instance  $p^{\text{new}}\in\Dtest$ can be obtained as $h_{\hat\varphi}(g(m^\text{new}), f(p^\text{new}))$.

The LLM-specific features $g(m_j)$ could include statistics on the training data of $m_j$ and architectural information (for example, number of attention layers and parameters). However, the high variety of hyperparameters involved in the definition and training of LLMs and the unavailability of detailed information on proprietary models makes defining broadly informative features hard, if not impossible. To circumvent this problem, we propose to use the performance of $m_j$ on a small set of reference instances $\Dref \subset \Dtrain$ as $g(m_j)$; in this way, it is sufficient to evaluate the new LLM $m^\text{new}$ on $\Dref$ to predict their performance on news instances $\Dtest$. See Figure~\ref{fig:teaser} for a graphical description of our method. Next, we discuss various methods to determine $\Dref$.

\subsubsection{Selecting the reference instances}
\label{sec:selector}
In order to select the most informative instances $(p_i, y_i) \in \Dtrain$ to form  $\Dref$ , we can use information intrinsic to the instances as well as the evaluation results of $\Ltrain$ on $\Dtrain$ (while keeping aside $\Dval$ and $\Lval$ to choose the best selection method; see Section~\ref{sec:choosing_best}). In general, let us denote by $x_i\in\R^d$ a feature vector associated to $p_i$ and $ \mathbf{X} \in \mathbb R^{d\times |\Dtrain|}$ the matrix whose columns are $x_i$. Finally, let us define $\mathbf Z^\text{train} = (z_{j,i})_{j:\  m_j\in\Ltrain, i:\ p_i\in\Dtrain}$.
We attempt using the following features:
\begin{itemize}
    \item features intrinsic to the prompt  $x_i =f(p_i)$, which are not necessarily the same used to build the assessor in Sections~\ref{sec:full_assessor} and \ref{sec:reference}; in this case, $d$ corresponds to the size of the features computed by $f$.
    \item     The binary successes/failure vector on $\Ltrain$, which identifies $\mathbf{X} = \mathbf{Z}^\text{train}$ and for which $d=\ntrain$.
    \item  The item demands obtained by applying the IRT approach in \cite{polo2024tinybenchmarks} (discussed in Section~\ref{sec:related}), which obtains a set of item demands and LLM capabilities starting from the success matrix $\mathbf Z^\text{train} $.
    Thus, we set $x_i$ to be the obtained item demands, whose size $d$ can be chosen by the user (we fix this to $d=10$ following \cite{polo2024tinybenchmarks}).
\end{itemize}

For all possible choices of $\mathbf X$ above, we apply the following methods to determine the reference instances:
\begin{itemize}
    \item  \textbf{Clustering on intrinsic features}: we perform KMeans clustering on  the columns of $\mathbf X$ and, for each identified cluster, add the instance $i$ closest to the cluster centroid to $\Dref$. The pre-specified number of clusters determines the number of selected instances.
    \item \textbf{Factor Analysis (FA)}: FA decomposes $\mathbf{X} = \mathbf W \mathbf{H}+\mathbf{E}$, where $\mathbf W \in \R^{d\times l}$ is the loading matrix, $\mathbf H\in\R^{l \times |\Dtrain|} $ is a matrix whose columns are the latent factors for each of the samples, $\mathbf E$ is a matrix of Gaussian noise and $l$ is the number of hidden factors. The features for each instance is assumed to be independent from the other instances given the matrix $\mathbf W$. In practice, we first fit FA with a high number of factors, set $l$ to the number of eigenvalues of the correlation matrix $\mathbf{X} \mathbf{X}^T$ which are larger than 1 and re-fit FA with the varimax rotation method \citep{kaiser1958varimax}. Then, we select the required number of reference instances by picking, for each $k=1, \ldots, l$, an approximately equal number of instances with the highest values of $|H_{k,i}|$\footnote{For instance, if we want to select 35 reference instances and $l=10$, we select the $i$'s corresponding to the top 4 $|H_{k,i}|$ for $k=1, \ldots, 5$ and those with the top 3 for $k=6, \ldots, 10$.}.

\end{itemize}
\looseness=-1
In total, we have 6 ways of selecting $\Dref$ (3 sets of features times 2 selection methods), two of which (clustering on success/failures and IRT item parameters) correspond to the selection method used in \cite{polo2024tinybenchmarks}. We compare these methods with a random reference subset; moreover, we also draw 20 random reference subsets, fit an assessor using the performance on the reference instances, and pick the random subset that leads to the highest performance (``random best of 20'').

\subsubsection{Predicting success by concatenating intrinsic features and performance on the reference instances}
\label{sec:reference_assessor}
Once we select the reference instances $\Dref$, we can extract the success of each LLM on $\Dref$ to define $g(m_j) = (z_{j,i})_{i\in\Dref}$. We can then concatenate this to the feature vector $f(p_i)$ (which does not need to be the same used for selecting the reference instances in Section~\ref{sec:selector}) and train a generic assessor aiming to minimise Eq.~\eqref{Eq:joint_assessor}. Notice how the features $f(p_i)$ can also rely on the reference dataset, as that is fixed for all new LLMs: for instance, we also attempt using a measure of similarity between the vector embeddings of $p_i$ and each of the instances in $\Dref$ as $f(p_i)$.%

\subsubsection{Choosing the best setup on validation data and predicting the performance of a new LLM}
\label{sec:choosing_best}
As mentioned above, we have multiple ways to define the reference set as well as multiple choices for the intrinsic features $f$. We can also choose multiple families of classifiers $h_\varphi$ and hyperparameters of the optimisation algorithm to minimise Eq.~\eqref{Eq:joint_assessor}. As such, we pick the combination of options which best predicts the performance of the validation LLMs $\Lval$ on the validation data $\Dval$. Hence, once we want to predict the performance of a new LLM $\mnew$ on a new instance $\xnew \in \Dtest$, we only need to evaluate $\mnew$ on $\Dref$ and apply the trained generic assessor. In our empirical studies below, we will test each method on multiple new LLMs, which we group into $\Ltest$.

\section{Empirical studies}
\label{sec:experiments}

\subsection{Considered datasets and splits}
\label{sec:datasets}

We consider two collections of datasets in our experiments\footnote{Code available at \url{https://github.com/LoryPack/ReferenceInstancesPredictability}}: 
\begin{itemize}
    \item \textbf{HELM-Lite} \cite{helm_lite}, a revised and reduced version of the popular HELM \citep{liang2022holistic}, which includes 10 different ``scenarios'' (i.e., datasets), some of which are stratified into sub-scenarios. Of those, we keep the scenarios and subscenarios for which the performance metric is binary, and further discard those for which different LLMs were tested with a different number of few-shot examples; the resulting subset spans 6 scenarios for a total of 4285 instances. The list of included and discarded scenarios and sub-scenarios can be found in Appendix~\ref{app:helm}. On this benchmark, the results for 30 LLMs from different families were available at the time we conducted our experiments (see Table~\ref{tab:llm_splits}).
    \item \textbf{KindsOfReasoning}, a collection that we introduce in this paper, which is aimed at testing various kinds of reasoning (logical, common sense, inductive, deductive, abductive, counterfactual, causal, analogical, spatial and arithmetic reasoning). Our collection includes 22 different datasets with varying number of instances, for a total of 37,529 instances. On this dataset, we tested all instruction-tuned models released from OpenAI, from \texttt{text-ada-001}\footnote{Note that the older models have been discontinued on 4th January 2024, but we obtained our raw results before that date.} to \\ \texttt{gpt-4-0125-preview}, for a total of 14 LLMs (see Table~\ref{tab:llm_splits}). The instance-level outputs of all models will be released\footnote{At \url{https://github.com/Kinds-of-Intelligence-CFI/KindsOfReasoning}}, in the spirit of \cite{burnell2023rethink}. To the best of our knowledge, this is the first collection of instance-level results covering all versions of a given model family from such a large time window, and we hope other researchers can find insights in this data. We provide more information about the construction of this collection in Appendix~\ref{app:kindsofreasoning}.
\end{itemize}

\begin{table*}[ht]
    \centering
    \caption{LLMs in $\Ltrain$, $\Lval$ and $\Ltest$ for the generic assessor experiments, on the two considered collection of datasets.}
    \begin{tabular}{lp{4.5cm}>{\raggedright\arraybackslash}p{10cm}}
\toprule
 & KindsOfReasoning & HELM-Lite \\
\midrule
Train & \texttt{openai/text-ada-001}, \texttt{openai/text-babbage-001}, \texttt{openai/text-curie-001}, \texttt{openai/text-davinci-001}, \texttt{openai/text-davinci-002}, \texttt{openai/gpt-3.5-turbo-0301}, \texttt{openai/gpt-3.5-turbo-0613}, \texttt{openai/gpt-3.5-turbo-1106} & \texttt{01-ai/yi-6b}, \texttt{01-ai/yi-34b}, \texttt{AlephAlpha/luminous-base}, \texttt{AlephAlpha/luminous-supreme}, \texttt{ai21/j2-grande}, \texttt{ai21/j2-jumbo}, \texttt{cohere/command}, \texttt{google/text-bison@001}, \texttt{google/text-unicorn@001}, \texttt{mistralai/mixtral-8x7b-32kseqlen}, \texttt{mistralai/mistral-7b-v0.1}, \texttt{openai/gpt-3.5-turbo-0613}, \texttt{openai/gpt-4-1106-preview}, \texttt{openai/text-davinci-002}, \texttt{openai/text-davinci-003}, \texttt{tiiuae/falcon-7b}, \texttt{writer/palmyra-x-v3}, \texttt{writer/palmyra-x-v2} \\
Validation & \texttt{openai/text-davinci-003}, \texttt{openai/gpt-3.5-turbo-0125} & \texttt{tiiuae/falcon-40b}, \texttt{openai/gpt-4-0613}, \texttt{AlephAlpha/luminous-extended}, \texttt{cohere/command-light} \\
Test & \texttt{openai/gpt-4-0125-preview}, \texttt{openai/gpt-4-0314}, \texttt{openai/gpt-4-0613}, \texttt{openai/gpt-4-1106-preview} & \texttt{anthropic/claude-2.1}, \texttt{anthropic/claude-2.0}, \texttt{anthropic/claude-instant-1.2}, \texttt{anthropic/claude-v1.3}, \texttt{meta/llama-2-70b}, \texttt{meta/llama-2-13b}, \texttt{meta/llama-2-7b}, \texttt{meta/llama-65b} \\
\bottomrule
\end{tabular}

    \label{tab:llm_splits}
\end{table*}

For each of these collections, we repeat all our experiments considering different choices for the train, validation, and test splits $\Dtrain, \Dval$ and $\Dtest$. In particular, we consider a random split, where the various splits are sampled by shuffling together all instances of all datasets. In addition, we consider multiple out-of-distribution (OOD) splits, where we keep one set of datasets as $\Dtest$ (according to some criteria), and $\Dtrain$ and $\Dval$ are obtained from randomly shuffling the other ones. In this way, the data used to train and select the best assessor (both in the generic and specific setup) have the same distribution, which is however different from the data where the selected assessor will be evaluated on. Details on the various splits are given in Table~\ref{tab:data_splits}.

\begin{table}[]
    \centering
    \caption{Size of $\Dtrain$, $\Dval$ and $\Dtest$ for the different splits for the KindsOfReasoning and HELM-Lite collections, together with the criteria for which datasets to include in the test split ($\Dtrain$ and $\Dval$ are randomly obtained from those not included in $\Dtest$).}
    \adjustbox{max width=\columnwidth}{%
    \begin{tabular}{lp{1cm}p{1cm}p{1cm}>{\raggedright\arraybackslash}p{2.5cm}}
\toprule
 & Train size & Validation size & Test size & Test set composition \\
\midrule
\multicolumn{5}{c}{\textit{KindsOfReasoning}} \\
\midrule
In-distribution & 21016 & 5254 & 11259 & Random \\
OOD 1 & 18069 & 4517 & 14943 & arithmetic \\
OOD 2 & 20705 & 5176 & 11648 & causal \\
OOD 3 & 21273 & 5318 & 10938 & logical, deductive, inductive, spatial, abductive, counterfactual, and analogical reasoning \\
OOD 4 & 23238 & 5810 & 8481 & world knowledge, common sense \\
\midrule
\multicolumn{5}{c}{\textit{HELM-Lite}} \\
\midrule
In-distribution & 2400 & 600 & 1285 & Random \\
OOD 1 & 2378 & 595 & 1312 & Math, GSM, MMU abstract algebra \\
OOD 2 & 2182 & 546 & 1557 & Legalbench \\
OOD 3 & 2295 & 574 & 1416 & Commonsense, Med QA, MMLU (except abstract algebra) \\
\bottomrule
\end{tabular}

    \label{tab:data_splits}
    }
\end{table}

Moreover, for the generic assessor experiments, we identify a single split of train, validation, and test LLMs $\Ltrain, \Lval$ and $\Ltest$ for each dataset collection. Analogously to how we selected the OOD splits, we make $\Ltest$ as different as possible from $\Ltrain$ and $\Lval$: concretely, we select LLMs from two producers as $\Ltest$ for HELM-Lite and all versions of \texttt{gpt4} models for KindsOfReasoning. In this way, we test the performance of our proposed methodology in the hard case where the new LLM we want to predict performance for is substantially different from the previously seen ones. The LLM splits are given in Table~\ref{tab:llm_splits}.

Notice how the diversity of LLMs in HELM-Lite is higher than that in KindsOfReasoning, as the latter has been evaluated on a single family of models. This is interesting as it allows us to understand how the performance of our proposed method changes when considering a broad or narrow set of LLMs.

\subsection{Considered prompt features}
Our methodology, discussed in Section~\ref{sec:methods}, relies on choosing a transformation $f$  that converts a given prompt $p_i$ into a set of numerical features $x_i = f(p_i)$, where we refer to these features as ``intrinsic'' as they do not depend on the particular LLM whose performance we are interested in predicting (as mentioned in Section~\ref{sec:reference_assessor}, in the generic assessor setup, we allow the intrinsic features to depend on the set of reference instances, as the set of reference instances is fixed for all LLMs in $\Ltest$). 

Empirically, we attempted using the following features:
\begin{itemize}
    \item the prompt embeddings computed from the OpenAI API (with the endpoint \texttt{text-embedding-3-large}, \cite{openai_embeddings});
    \item the \texttt{Word2vec} \cite{Mikolov2013EfficientEO} and \texttt{FastText} \cite{bojanowski2017enriching} word embeddings, which compute a vector for each word of the prompt which we average to obtain a feature vector for the whole prompt;
    \item the 1-gram vectors, which are obtained as a measure of the frequency of words in a specific prompt normalized over that of the words in the entire set of training prompts.
\end{itemize}

We studied the performance of these in the specific assessor setup (complete results in Appendix~\ref{app:results}) and found that OpenAI embeddings perform better more frequently. Moreover, the OpenAI embeddings obtained from the endpoint \texttt{text-embedding-3-large} were trained using Matryoshka Representation Learning \cite{kusupati2022matryoshka}, which allows them to be truncated (by removing the final elements of the vector) without the embedding losing its concept-representing properties. As such, we investigate the performance of the specific assessor by truncating the OpenAI embeddings (Appendix~\ref{app:n_emb}) and we found that the performance saturates using 1024 (out of a total of 3072) embeddings. As such, our experiments on generic assessors use the first 1024 elements of the OpenAI embeddings. Additionally, in the generic assessor framework, we also attempt to use the cosine similarity between the embeddings of each element of the selected $\Dref$ and the considered instance $p_i$ as $f(p_i)$. 

\subsection{Metrics and other details}
\label{sec:metric}
\looseness=-1
We use the Area Under the Curve (AUC) as a metric for the performance of the generic and specific assessor. The AUC measures how well a binary probabilistic classifier (i.e., a classifier that provides a probabilistic prediction for a binary variable) discriminates between the two classes: a classifier whose assigned probabilities for the two classes do not overlap achieves the maximum value $\text{AUC} = 1$, while a classifier assigning random values to the two classes achieves $\text{AUC}=0.5$. We employ the AUC as its extreme values are insensitive to the proportion of positive and negative samples in the dataset, and it can therefore be used to compare results across various scenarios (in our case, the various train/validation/test splits and the two dataset collections). However, AUC is insensitive to monotonic transformation of the output probabilities and this implies that a classifier achieving $\text{AUC}=1$ can be miscalibrated (for instance, a classifier assigning probability 0.51 to all positive samples and 0.49 to all negative samples achieves $\text{AUC}=1$, but its predictions are miscalibrated).

We test various values of the size of $\Dref$ 
(results in Appendix~\ref{app:n_ref}) and we find that the performance on the test set saturates around 100 reference instances; as such, all results reported in the main text are obtained with that value.

Next, for any data split and any choice of $\Dref$ in the generic assessor setup, we attempt to use various classifiers as assessors (logistic regression with $l_2$ and $l_1$ penalty and \texttt{xgboost}). Furthermore, as mentioned in Section~\ref{sec:reference_assessor}, in the generic assessor setup, we attempt to use the OpenAI embeddings as well as their cosine similarity to those of the elements of $\Dref$ as instance features. To select the best method, we do the following: 
\begin{itemize}
    \item in the specific assessor setup, we compute the AUC of each classifier trained on each test LLM on $\Dval$, pick the one with the highest value, and report the AUC of that classifier on $\Dtest$.
    \item In the generic assessor setup, for each data split, we evaluate the AUC of each combination of classifier, choice of $\Dref$ and instance features $f$ on $\Dval$ for each LLM in $\Lval$. To select the best combination, we compute the win rate of each combination for each validation LLM and pick the combination with the highest average win rate over $\Lval$ (a simpler average over $\Lval$ is impacted by the intrinsic different predictability of the different LLMs, which change the maximally achievable AUC).
\end{itemize}

Notice how, by doing so, the specific assessor requires each test LLM to be tested on the whole $\Dtrain$ and $\Dval$, while the generic assessor only uses the results of $\Ltrain$ and $\Lval$ on $\Dtrain$ and $\Dval$ correspondingly and requires evaluating each test LLM on $\Dref$. The latter case is therefore fully representative of the case of a new LLM evaluated on a new instance. The winning combination for each data split is reported in Table~\ref{tab:best_combinations}. Interestingly, for multiple data splits, the randomly sampled $\Dref$ performs better than those determined according to the advanced criteria in Section~\ref{sec:selector}. While surprising at first, other works \cite{ye2023how, wang2023challenging, polo2024tinybenchmarks, kipnis2024metabench} had found that benchmarks can be reduced by random sampling for multiple purposes.

In terms of classifier, instead, XGBoost generally performs better. Finally, using similarity between the embeddings of the reference instances and those of the considered instance more frequently performs better than directly using the latter as $f(p)$.

\begin{table*}
    \centering
    \caption{The best combination of instance-intrinsic features, selector and classifier for each data split in the two considered dataset collections, selected according to the performance on validation LLMs as discussed in Section~\ref{sec:metric}. 
    In the ``instance-intrinsic features'' column, ``embeddings'' refers to using the OpenAI embeddings of the considered instance as $f(p_i)$, while ``similarity'' refers to using the cosine similarity between the OpenAI embeddings of the reference instances and that of the considered instance; further, ``similarity with interaction'' explicitly adds features obtained as the pairwise produce of each similarity with its corresponding success (notice that this is superfluous for XGBoost, which can natively leverage interactions between features).
    }
    \label{tab:best_combinations}
    \begin{tabular}{llll}
\toprule
 & Instance-intrinsic features & Selector & Classifier \\
\midrule
\multicolumn{4}{c}{\textit{KindsOfReasoning}} \\
\midrule
In-distribution & Similarity & Random best of 20 & XGBoost \\
OOD 1 & Similarity & Factor analysis embeddings & XGBoost \\
OOD 2 & Similarity with interaction & Clustering IRT values & XGBoost \\
OOD 3 & Embeddings & Random & XGBoost \\
OOD 4 & Similarity & Random & XGBoost \\
\midrule
\multicolumn{4}{c}{\textit{HELM-Lite}} \\
\midrule
In-distribution & Similarity with interaction & Clustering embeddings & Logistic Regression L1 C=0.1 \\
OOD 1 & Embeddings & Clustering LLM success & XGBoost \\
OOD 2 & Similarity with interaction & Random & Logistic Regression L1 C=1 \\
OOD 3 & Similarity with interaction & Clustering LLM success & Logistic Regression L1 C=1 \\
\bottomrule
\end{tabular}

\end{table*}

\subsection{How well can we predict success?}
\label{sec:experiments_performance}

\begin{figure*}
    \centering
    \includegraphics[width=1\textwidth]{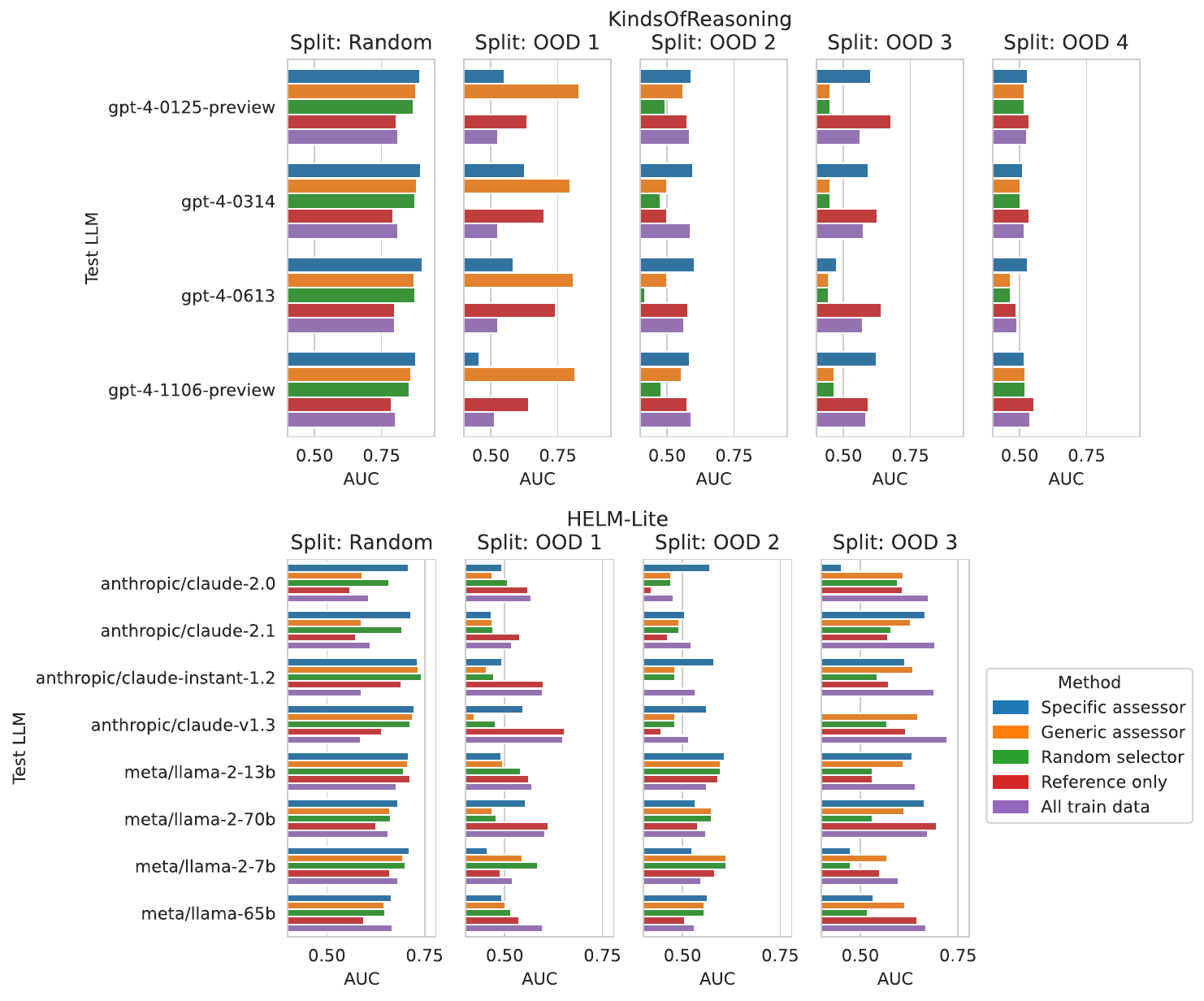}      
    \caption{Predictive performance (AUC) of the specific and generic assessor and a few baselines, for different splits of the KindsOfReasoning and HELM-Lite collections of datasets. Some combinations (for instance, the random selector on split 1 of KindsOfReasoning achieve AUC lower than the lower bound of the panels (0.4) and are hence hidden in the graph.}
    \label{fig:main}
\end{figure*}

Figure~\ref{fig:main} includes our main result, namely the predictive performance of the generic and specific assessor for the test LLMs $\Ltest$, alongside three baselines:
\begin{itemize}
    \item ``Random selector'' corresponds to a generic assessor where $\Dref$ is a random set of instances selected from $\Dtrain$, instead of using the selection methods discussed in Section~\ref{sec:selector}; the best classifier and intrinsic features for each split are chosen using validation data and LLMs as for the generic assessor (Section~\ref{sec:metric}). 
    \item ``Reference only'' is obtained by fitting, for each $\Ltest$, an assessor on the performance on the elements of $\Dref$, by only taking as input the intrinsic features of the prompts in $\Dref$ (thus, ignoring the performance of the previous LLMs to predict the performance of the new one). Notice how the best classifier and $\Dref$ for ``reference only'' are selected independently of those for the generic assessor by using the validation data and LLMs as discussed in Section~\ref{sec:metric}.
    \item ``All train data'' is obtained by fitting a single assessor on the pooled performance results of all LLMs in $\Ltrain$ on $\Dtrain$ only using the intrinsic features $f(p_i)$ (effectively considering all LLMs as a single LLM and ignoring the new LLM's performance on $\Dref$)
\end{itemize}

From the results in Figure~\ref{fig:main}, several considerations can be made. First, notice how the predictive performance generally degrades out of distribution with respect to the in-distribution (random) split. For some out-of-distribution splits, some predictive power remains (recall that $\text{AUC}=0.5$ corresponds to random guess) but, on other splits, even the specific assessor performs poorly, despite relying on evaluation results of the test LLMs on the whole train and validation data splits. This indicates that the considered intrinsic features of the prompt (the OpenAI embeddings) do not reliably capture a general performance pattern. While, in principle, more informative features could be used, it is also possible that there is an inner limit to the out-of-distribution predictability of the current generation of LLMs, due to their stochastic nature.

Moreover, the specific assessor always outperforms our generic assessor in distribution and does so frequently out of distribution, as expected from the former having access to more information about the test LLM; however, the performance gap is generally small. In distribution, further, the generic assessor almost always outperforms or performs comparably with the ``all train data'' and ``reference only''  baselines, indicating that combining the information on previous LLMs and the evaluation results of the test LLM on $\Dref$ generally performs better than relying only on either one. For some OOD splits (OOD 2 and 3 for KindsOfReasoning and OOD1 and 3 for HELM-Lite), instead, either or both of these baselines perform better than the generic assessor, indicating how the generic assessor likely overfits to the training distribution; however, in most of those cases, the predictive performance is quite low for all methods (except for split 3 in HELM-Lite). 

\looseness=-1
If we instead compare the generic assessor with the ``random selector'' baseline (which is identical to the generic assessor but with a random $\Dref$), we see how the two often perform comparably and there are a few cases where either one prevails, in roughly equal frequency. This indicates that the generic assessor is not sensitive to the specific selection of $\Dref$ (an indication for this could also be seen in Table~\ref{tab:best_combinations}, where there is no coherent best selector and where a few times the ``random'' subset was selected as best). Notice how, on validation data, the selected combination of selector, features, and classifier for the generic assessor is always better than the random selector baseline, as the possible choices for the latter are a subset of those for the former; however, our Figure~\ref{fig:main} shows how, at least in a few cases, it is possible that the random selector performs better on test data.

In a similar manner, the ``reference only'' baseline is identical to a ''specific assessor'' trained on a subset of $\Dtrain$, but with the selection of the best classifier being carried out on the validation LLMs, instead of using the results of the considered LLM on $\Dval$. Still, the specific assessor always performs better than ``reference only'' in-distribution, while the latter sometimes overtakes the former out-of-distribution, indicating that the specific assessor overfits the training distribution due to the larger number of training points or due to the classifier selection being performed using the test LLM.

\section{Conclusion}
\label{sec:conclusion}

\looseness=-1
We proposed a novel framework for predicting the performance of a new Large Language Model (LLM) on individual task instances by leveraging the evaluation results of previously tested LLMs. Our approach minimises the number of evaluations required for a new LLM by introducing a \textit{generic assessor} that combines instance-specific features with LLM-specific features derived from performance on a small set of reference instances. In doing so, our method is cheaper than specific assessors \cite{zhou2022reject, hernandez2022training} that solely rely on the performance of the considered LLM on a larger set of labelled examples. While we focus on LLMs, our methodology can be seamlessly applied to predict the performance of other AI systems, by using suitable system-specific and instance-specific features.

\looseness=-1
We conducted empirical studies on the HELM-Lite and KindsOfReasoning collections. In distribution, we found our generic assessor to perform only slightly worse than the specific assessor, indicating that the generic assessor is a viable method to reduce the evaluation cost of new LLMs when interested in predicting instance-level performance. In distribution, moreover, the generic assessor almost always outperforms the baseline relying only on information on previous LLMs or on the results of the test LLM on $\Dref$. Further, the generic assessor is mostly unsensitive to the specific set of reference instances used. Out of distribution, instead, the picture is more varied: no clear winner emerges, and instance-level predictability is generally low, except for a few splits (for instance, OOD 1 in KindsOfReasoning and OOD 3 in HELM-Lite). 
As such, our work raises awareness of the low inner predictability of LLMs, and we hope it encourages the research community to focus more on characterising what affects the predictability of LLMs and hence finding ways to increase it, which will help to make AI systems more reliable \cite{zhou2023predictable}.
To foster research in this area, we release the instance-level results of all instruction-finetuned GPT3 and GPT4 models until \texttt{gpt4-0125-preview} on our novel KindsOfReasoning collection of datasets; to the best of our knowledge, this is the first publicly available set of fine-grained results for all versions of an LLM family.

\label{sec:limitations}

Our work has several limitations, which can be addressed in future work: 
\begin{itemize}
    \item First, as we focused on providing a proof of concept, it is possible that optimised features and bespoke classifiers can better predict instance-level performance of LLMs.
    \item Moreover, it may be possible to identify other LLM-specific behavioural features more suitable than the performance on a reference set of instances we employed. In particular, it may be possible to adaptively select the most informative features (such as done in \cite{kipnis2024metabench}) and thus improving accuracy while reducing the number of required evaluations even more. 
    \item \looseness=-1 Finally, in our out-of-distribution studies, we considered the same data distribution for the train and validation split, and a different one for the test split. With this setup, we found our generic assessor sometimes underperforms the baselines on the test data, likely due to overfitting the training distribution. This may be prevented by using validation data with a different distribution from the train (and test) split.
\end{itemize}

\begin{acks}
Lorenzo Pacchiardi is funded by US DARPA grant HR00112120007 (RECoG-AI), received computational support from OpenAI through the Researcher Access Program and travel support from TAILOR Connectivity Fund (EU’s Horizon 2020 research and innovation programme, grant agreement No. 952215). The authors thank Marko Tesic and Lexin Zhou for their useful feedback on the draft.
\end{acks}

\bibliographystyle{ACM-Reference-Format}
\bibliography{bibliography}

\appendix

\section{Appendix / supplemental material}

\subsection{More information on the considered and excluded scenarios from HELM-Lite}
\label{app:helm}
As mentioned in the main text, we discard some scenarios and subscenarios from HELM-Lite as either the performance metric was non-binary or because the available results used a different number of few-shot prompts for different LLMs. In particular, we discard the following:
\begin{itemize}
    \item LegalBench:
    \begin{itemize}
        \item corporate lobbying - incoherent number of few-shots across LLMs
    \end{itemize}
    \item MATH:
    \begin{itemize}
        \item algebra - incoherent number of few-shots across LLMs
        \item geometry - incoherent number of few-shots across LLMs
        \item intermediate algebra - incoherent number of few-shots across LLMs
    \end{itemize}
    \item NarrativeQA: non-binary metric (f1 score)
    \item NaturalQuestions: non-binary metric (f1 score)
    \item WMT 2014: non-binary metric (BLEU score)
\end{itemize}

As such, the subset of HELM-Lite that we consider throughout our experiments is made up of the following scenarios and subscenarios:
\begin{itemize}
    \item commonsense
    \item GSM8K
    \item MedQA
    \item LegalBench:
    \begin{itemize}
        \item abercrombie
        \item function of decision section
        \item proa
        \item international citizenship questions
    \end{itemize}
    \item MATH:
    \begin{itemize}
        \item counting and probability
        \item number theory
        \item prealgebra
        \item precalculus
    \end{itemize}
    \item MMLU:
    \begin{itemize}
        \item abstract algebra
        \item college chemistry
        \item computer security
        \item econometrics
        \item US foreign policy
    \end{itemize}
\end{itemize}

\subsection{The KindsOfReasoning collection}
\label{app:kindsofreasoning}

Table~\ref{tab:kindsofreasoning} shows detailed information on the datasets included in the KindsOfReasoning collection. For some datasets, we only kept a smaller number of instances than the one available, to reduce the cost of evaluating a model on the full benchmark. We do not do this for the ``Arithmetic'' dataset as each of the prompt of that dataset is short, and hence the cost of evaluating it is small (besides, we use Arithmetic as the test data for one of our chosen splits, and subsampling it would have made the test data too small).

Most of the datasets included in this collection are present in one (or more) of BIG-Bench \cite{srivastava2022beyond}, LogiGLUE \cite{luo2023towards}, CALM-bench \cite{dalal2023calm} and GLoRE \cite{teng2023glore}. However, as mentioned in the main text~\ref{sec:related}, our collection covers more kinds of reasoning. The dataset and the instance-level results of all instruct-GPT models from OpenAI (from\texttt{text-ada-001} to \texttt{gpt4-0125-preview} will be released at \texttt{anonymised}).

\begin{table*}%
\caption{Datasets used in building the KindsOfReasoning collection. See Appendix~\ref{app:kindsofreasoning} for information on the column meanings.
}
\label{tab:kindsofreasoning}
\adjustbox{max width=\textwidth}{%
    \begin{tabular}{@{}p{2.5cm}p{2.5cm}p{1.5cm}p{2cm}p{1cm}p{1cm}p{1cm}p{2.5cm}p{1.5cm}@{}}
\toprule
\textbf{Task name} & \textbf{Reasoning type} & \textbf{Used in} & \textbf{Task Type} & \textbf{Used split} & \textbf{N samples} & \textbf{N samples used} & \textbf{Notes} & \textbf{Source used} \\ \midrule
formal fallacies\newline syllogisms negation \cite{srivastava2022beyond} & Logical reasoning & BIG-Bench  & Valid/invalid & - & 14200 & 1000 & - & BIG-Bench  \\
logical\_args \cite{srivastava2022beyond}& Logical reasoning \newline common sense & BIG-Bench  & MC (5) & - & 32 & 32 & - & BIG-Bench  \\
babi\_task\_16 \cite{srivastava2022beyond}& inductive reasoning & LogiGLUE  & 1-word answer & test & 5000 & 1000 & - & BIG-Bench  \\
LogiQA 2.0 \cite{LogiQA}  & deductive reasoning & {LogiGLUE \newline GLoRE}  & MC (4) & validation & 1569 & 1569 &\footnotemark{}  & \href{https://github.com/openai/evals/pull/648}{OpenAI \texttt{evals} library}  \\ 
wanli \cite{wanli} & deductive reasoning & LogiGLUE  & NLI & test & 5000 & 1000 & Slightly modified the prefix & \href{https://huggingface.co/datasets/logicreasoning/logi_glue}{LogiGLUE}  \\
alpha\_nli \cite{alphanli} & abductive & {CALM-bench\newline LogiGLUE}  & MC (2) & test & 1432 & 1000 & Changed from NLI to MC format & \href{https://huggingface.co/datasets/logicreasoning/logi_glue}{LogiGLUE}   \\
reclor \cite{ReClor} & {abductive, inductive,\newline deductive reasoning} & LogiGLUE\newline GLoRE  & MC (4 options) & test & 500 & 500 &\footnotemark{}  & \href{https://github.com/openai/evals/pull/648}{OpenAI \texttt{evals} library}  \\
crass\_ai \cite{srivastava2022beyond} & Counterfactual reasoning & BIG-Bench  & MC (5 options) & - & 44 & 44 & - & BIG-Bench  \\
cause and effect \cite{srivastava2022beyond}& Causal reasoning & BIG-Bench  & MC (2) & - & 102 & 102 & Over 2 different formats & BIG-Bench  \\
fantasy reasoning  \cite{srivastava2022beyond}& Causal reasoning & BIG-Bench  & Yes/No & - & 201 & 201 & - & BIG-Bench  \\
goal step inference \cite{srivastava2022beyond}& Causal reasoning & BIG-Bench  & MC (4) & - & 7053 & 3000 & Over 3 subtasks & BIG-Bench  \\
Copa \cite{copa} & Causal reasoning,\newline world knowledge & CALM-bench  & MC (2) & test & 500 & 500 & - & \href{https://people.ict.usc.edu/~gordon/downloads/COPA-resources.tgz}{Original source}  \\
Cosmos\_qa \cite{cosmos_qa}  & Causal reasoning,\newline world knowledge & CALM-bench  & MC (4) & validation & 2985 & 2985 & use validation set as the test set does not have labels. & \href{https://huggingface.co/datasets/cosmos_qa}{HuggingFace} \\
ropes\cite{Ropes} & Causal reasoning,\newline world knowledge & CALM-bench  & Completion & validation & 1688 & 1688 & use validation set as the test set does not have labels. & \href{https://huggingface.co/datasets/ropes}{HuggingFace}   \\
Anli \cite{anli} & Causal reasoning,\newline world knowledge & LogiGLUE  & NLI & test & 3200 & 3200 & Merged the 3 “rounds” (levels of difficulty) together & \href{https://dl.fbaipublicfiles.com/anli/anli_v1.0.zip}{Original source}   \\
Emoji\_movie \cite{srivastava2022beyond} & analogical reasoning, world knowledge & BIG-Bench  & MC (5) & - & 100 & 100 & - & BIG-Bench  \\
abstract narrative\newline understanding \cite{srivastava2022beyond} & analogical reasoning & BIG-Bench  & MC (10 and 100) & - & 2000 & 2000 & Over 2 subtasks (9 and 99 distractors; I discarded the one with 4 distractors) & BIG-Bench  \\
odd one out \cite{srivastava2022beyond} & analogical reasoning & BIG-Bench  & MC (variable number) & - & 86 & 86 & - & BIG-Bench  \\
metaphor understanding \cite{srivastava2022beyond} & analogical reasoning & BIG-Bench  & True/False & - & 680 & 680 & - & BIG-Bench  \\
geometric shapes \cite{srivastava2022beyond} & Spatial reasoning & BIG-Bench  & MC (10) & - & 360 & 360 & - & BIG-Bench  \\
Space\_nli \cite{spaceNLI} & Spatial reasoning & - & NLI & - & 1604 & 1604 & - & \href{https://github.com/kovvalsky/SpaceNLI}{Original source}   \\
Arithmetic \cite{srivastava2022beyond} & Arithmetic ability & BIG-Bench  & Completion & - & 15023 & 15023 & Over 20 subtasks & BIG-Bench  \\ \bottomrule
\end{tabular}
}
\end{table*}

\footnotetext[7]{I use the multiple-choice version rather than the NLI one; moreover, the source I used shuffled the order of options and replaced the correct option with “none is correct”, so the model should always select that.} %

\footnotetext{The source I used shuffled the order of options and replaced the correct option with “none is correct”, so the model should always select that.}

\subsection{Additional results with other features in the specific assessor setup}
\label{app:results}

\begin{figure*}
    \centering
    \begin{subfigure}[b]{\textwidth}
        \includegraphics[width=\textwidth]{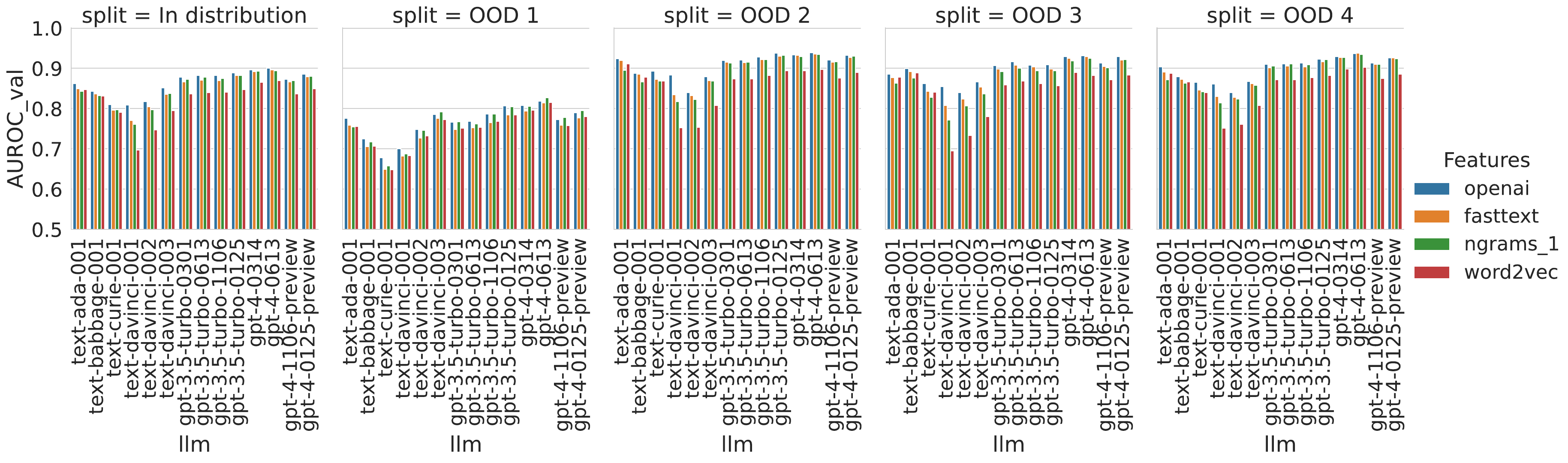}
        \caption{AUC with different choices of instance-intrinsic features on $\Dval$.}        
    \end{subfigure}
    \begin{subfigure}[b]{\textwidth}
        \includegraphics[width=\textwidth]{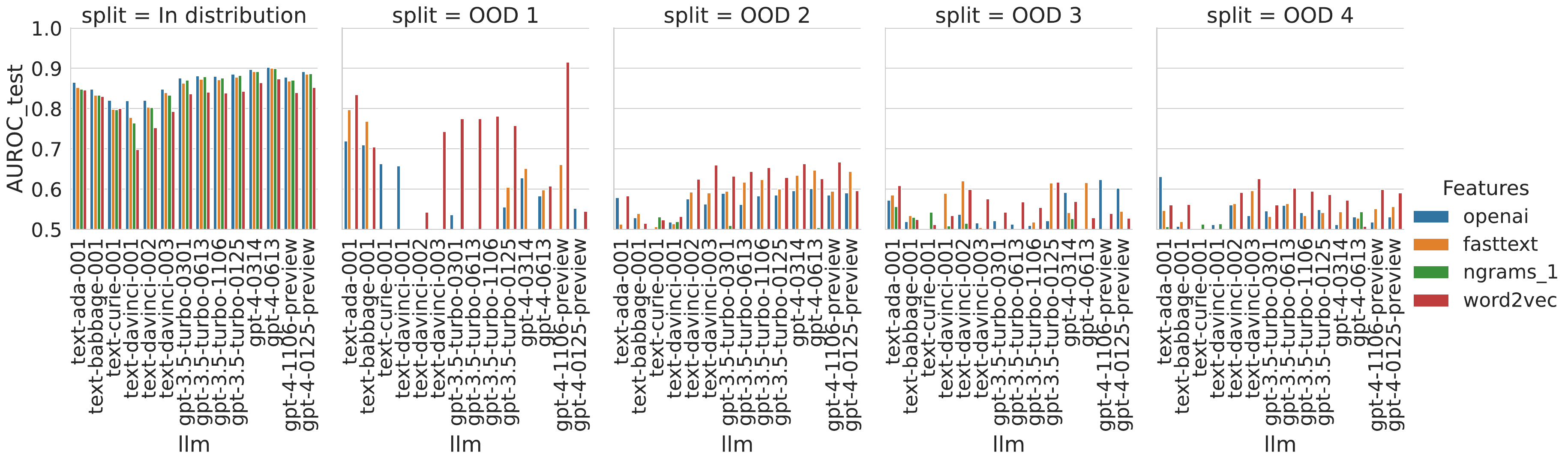}
        \caption{AUC with different choices of instance-intrinsic features on $\Dtest$.}        
    \end{subfigure}
    \caption{AUC with different choices of instance-intrinsic features (OpenAI embeddings, Word2Vec, FastText and 1-gram), for different splits on KindsOfReasoning. For each split and feature, various classifiers were trained on $\Dtrain$ and the best according to its performance on $\Dval$ was selected; the panels report the performance of the latter on $\Dval$ and $\Dtest$.}
    \label{fig:other_features_reasoning}
\end{figure*}

\begin{figure*}
    \centering
    \begin{subfigure}[b]{\textwidth}
        \includegraphics[width=\textwidth]{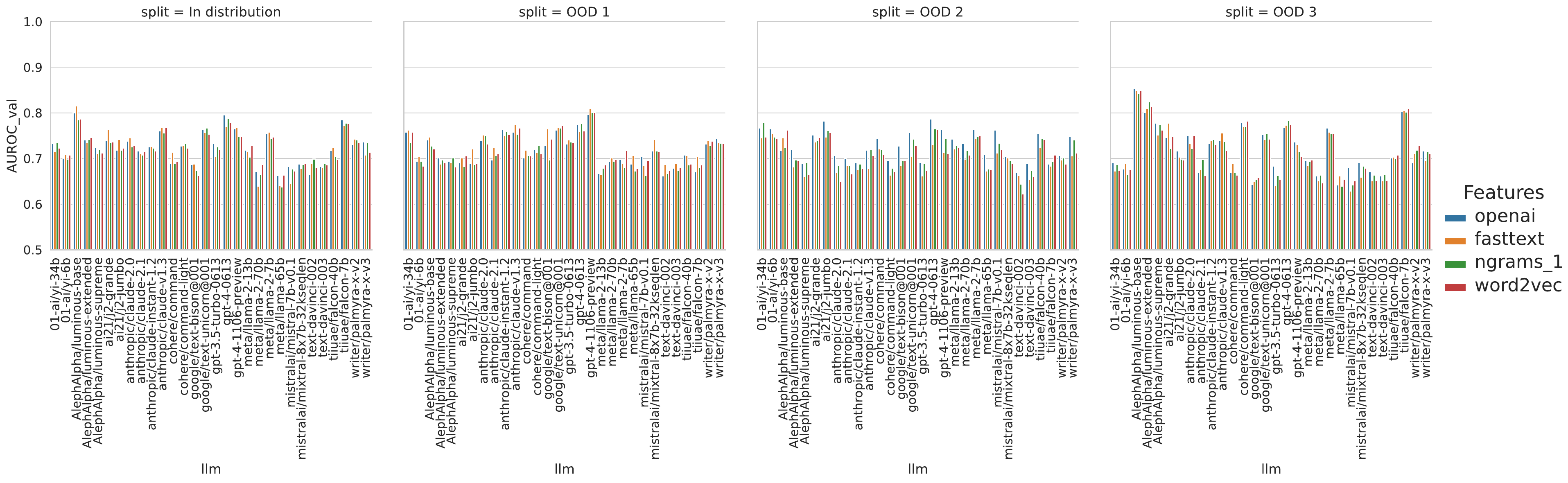}
        \caption{AUC with different choices of instance-intrinsic features on $\Dval$.}        
    \end{subfigure}
    \begin{subfigure}[b]{\textwidth}
        \includegraphics[width=\textwidth]{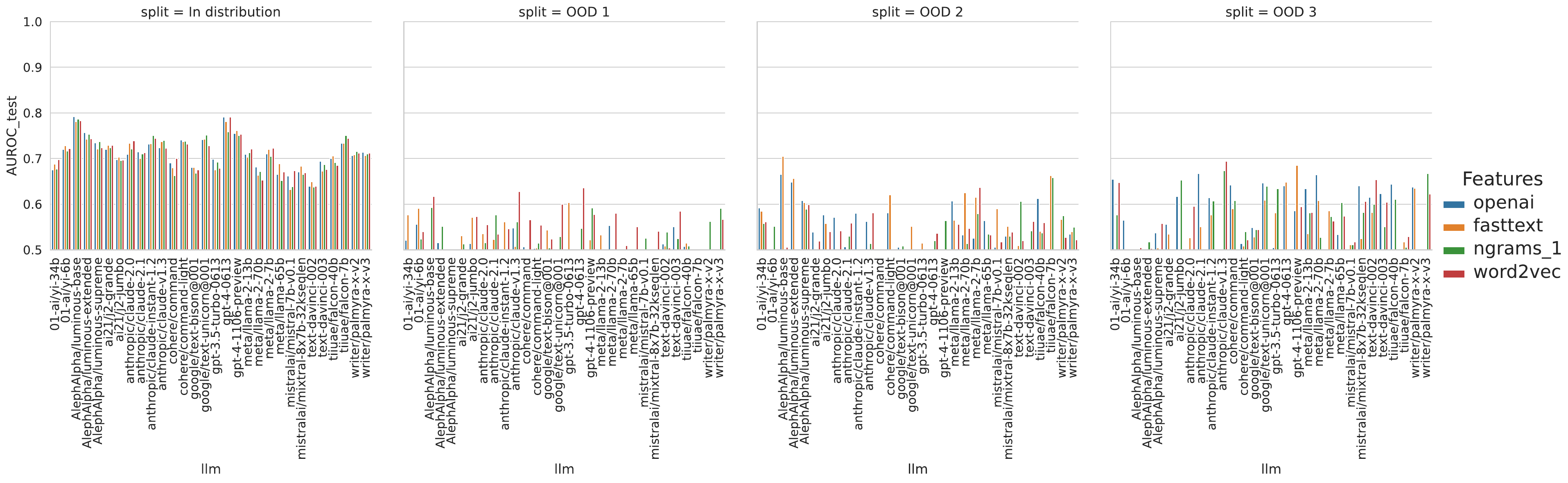}
        \caption{AUC with different choices of instance-intrinsic features on $\Dtest$.}        
    \end{subfigure}
    \caption{AUC with different choices of instance-intrinsic features (OpenAI embeddings, Word2Vec, FastText and 1-gram), for different splits on HELM-Lite. For each split and feature, various classifiers were trained on $\Dtrain$ and the best according to its performance on $\Dval$ was selected; the panels report the performance of the latter on $\Dval$ and $\Dtest$.}
    \label{fig:other_features_helm}
\end{figure*}

Figures~\ref{fig:other_features_reasoning} and \ref{fig:other_features_helm} show performance of the specific assessor setup using different features intrinsic to the prompt, for different data splits of the KindsOfReasoning and HELM-Lite collections respectively. In particular, for each figure, the top panel shows performance on $\Dval$, while the latter shows performance on $\Dtest$, for the classifier selected according to its best performance on $\Dval$. On the validation data, the performance of the OpenAI embeddings is generally higher and, as such, the experiments reported in the main text are with this choice of embeddings. However, the performance on $\Dtest$ for the OOD splits show a mixed picture, with the OpenAI embeddings often performing worse than simpler ones (such as Word2Vec) and with generally lower performance.

\subsection{How many OpenAI embeddings are needed?}
\label{app:n_emb}

\begin{figure*}
    \centering
    \begin{subfigure}[b]{\textwidth}
        \includegraphics[width=\textwidth]{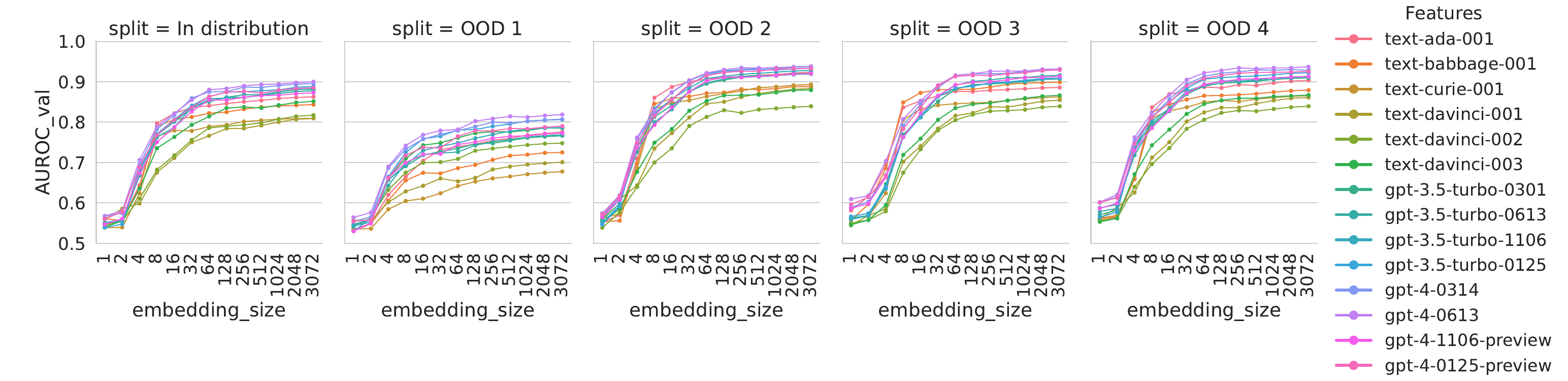}
        \caption{AUC with increasing number of OpenAI embeddings on $\Dval$.}        
    \end{subfigure}
    \begin{subfigure}[b]{\textwidth}
        \includegraphics[width=\textwidth]{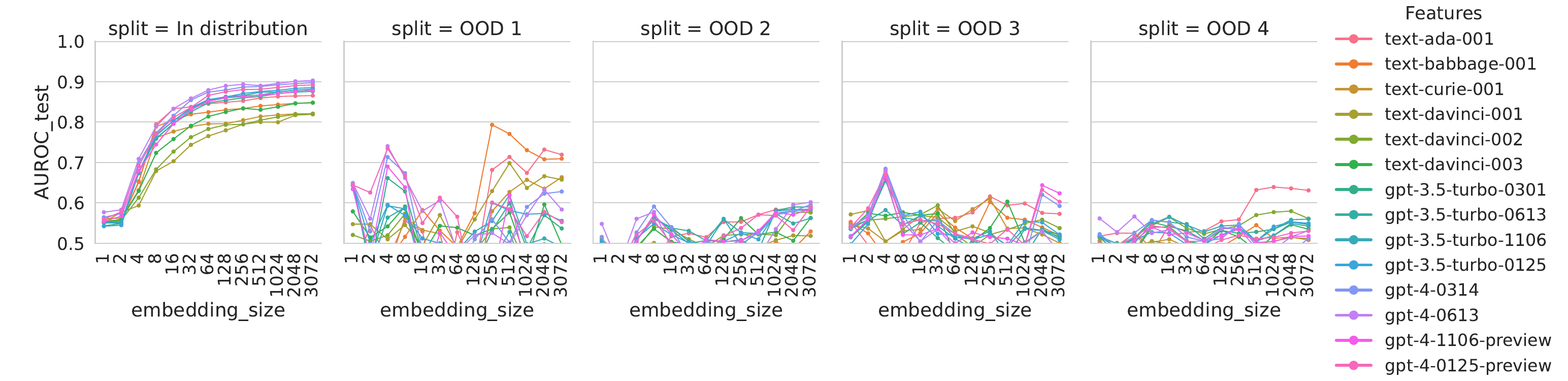}
        \caption{AUC with increasing number of OpenAI embeddings on $\Dtest$.}        
    \end{subfigure}
    \caption{AUC with increasing number of OpenAI embeddings for specific assessors trained on increasing number of OpenAI embeddings, for different splits on KindsOfReasoning. For each split and number of embeddings, various classifiers were trained on $\Dtrain$ and the best according to its performance on $\Dval$ was selected; the panels report the performance of the latter on $\Dval$ and $\Dtest$.}
    \label{fig:n_emb_reasoning}
\end{figure*}

\begin{figure*}
    \centering
    \begin{subfigure}[b]{\textwidth}
        \includegraphics[width=\textwidth]{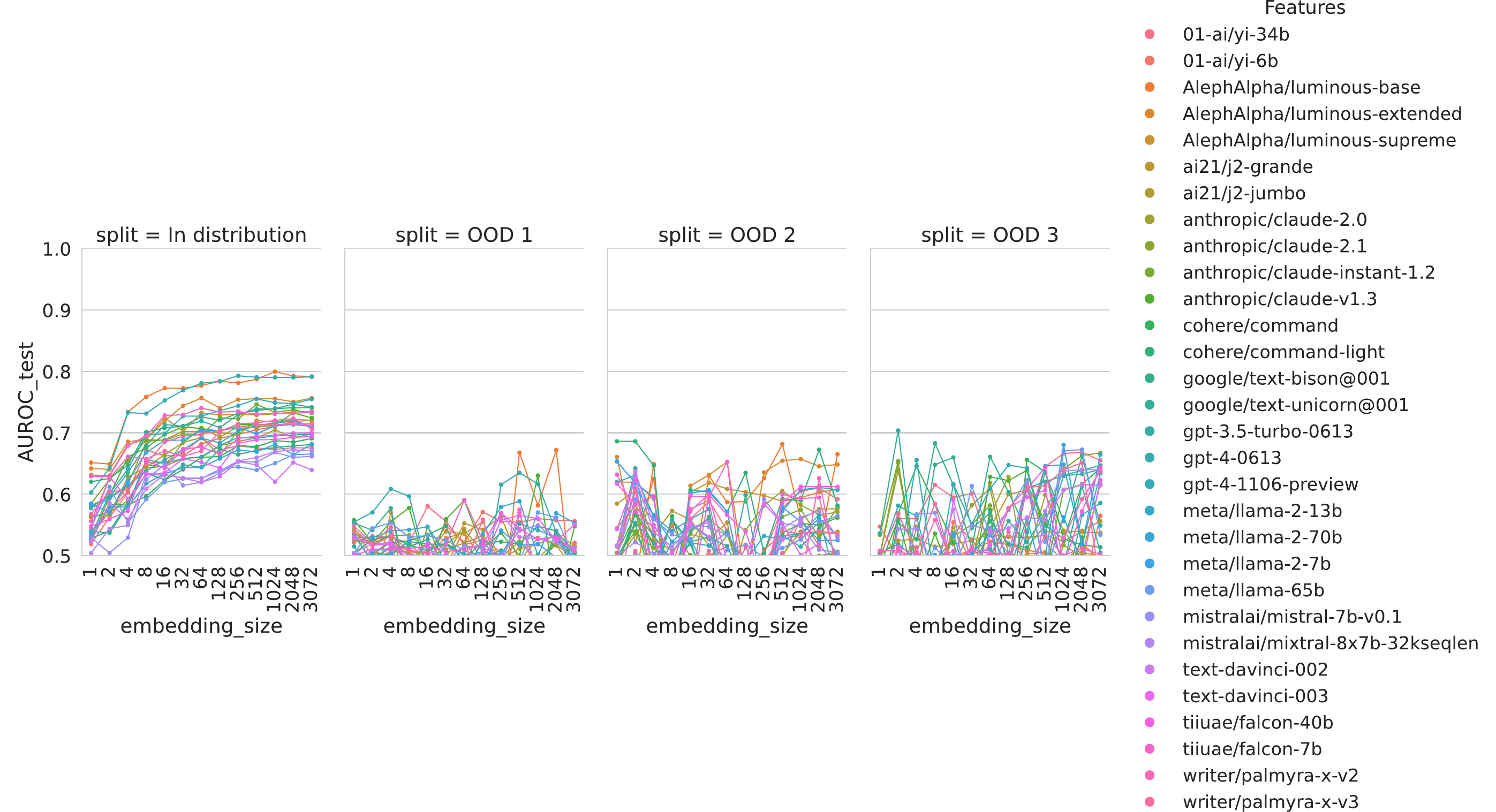}
        \caption{AUC with increasing number of OpenAI embeddings on $\Dval$.}        
    \end{subfigure}
    \begin{subfigure}[b]{\textwidth}
        \includegraphics[width=\textwidth]{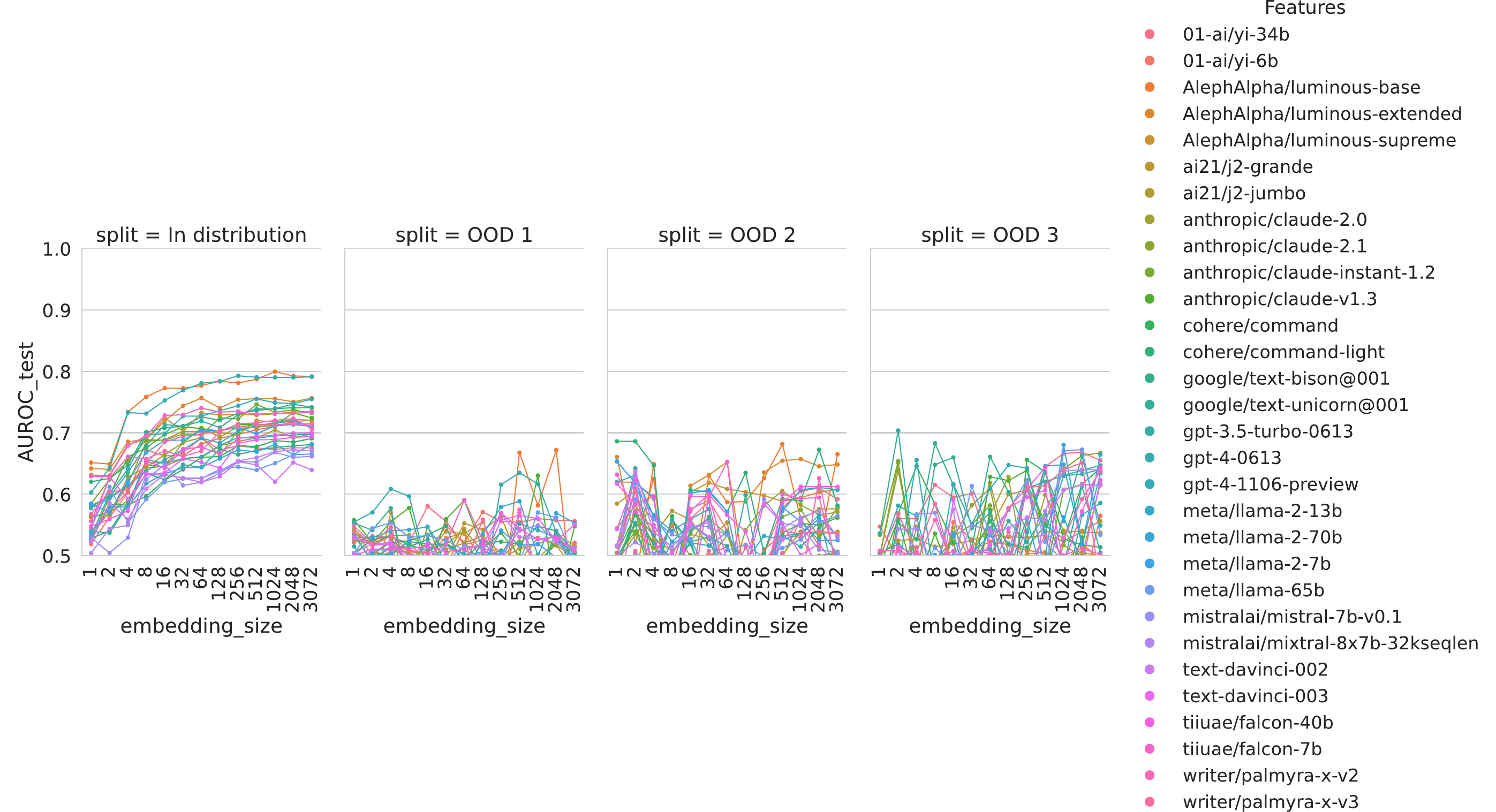}
        \caption{AUC with increasing number of OpenAI embeddings on $\Dtest$.}        
    \end{subfigure}
    \caption{AUC with increasing number of OpenAI embeddings for specific assessors trained on increasing number of OpenAI embeddings, for different splits on HELM-Lite. For each split and number of embeddings, various classifiers were trained on $\Dtrain$ and the best according to its performance on $\Dval$ was selected; the panels report the performance of the latter on $\Dval$ and $\Dtest$.}
    \label{fig:n_emb_helm}
\end{figure*}

Figures~\ref{fig:n_emb_reasoning} and \ref{fig:n_emb_helm} show performance of the specific assessor using the OpenAI embeddings truncated at different vector sizes, for different data splits of the KindsOfReasoning and HELM-Lite collections respectively. In particular, for each figure, the top panel shows performance on $\Dval$, while the latter shows performance on $\Dtest$, for the classifier selected according to its best performance on $\Dval$. The performance on $\Dval$ (and $\Dtest$ for the in-distribution split) plateaus when the truncation size reaches $1024$ and, as such, all the results reported in the main text are with that truncation size.  On $\Dtest$ for the various OOD splits, the performance does not follow a smooth curve, but still seems to peak more often around a truncation size of 1024.

\subsection{Control for number of training samples in the KindsOfReasoning collection}
\label{app:training_samples}

\begin{figure*}
    \centering
    \begin{subfigure}[b]{\textwidth}
        \includegraphics[width=\textwidth]{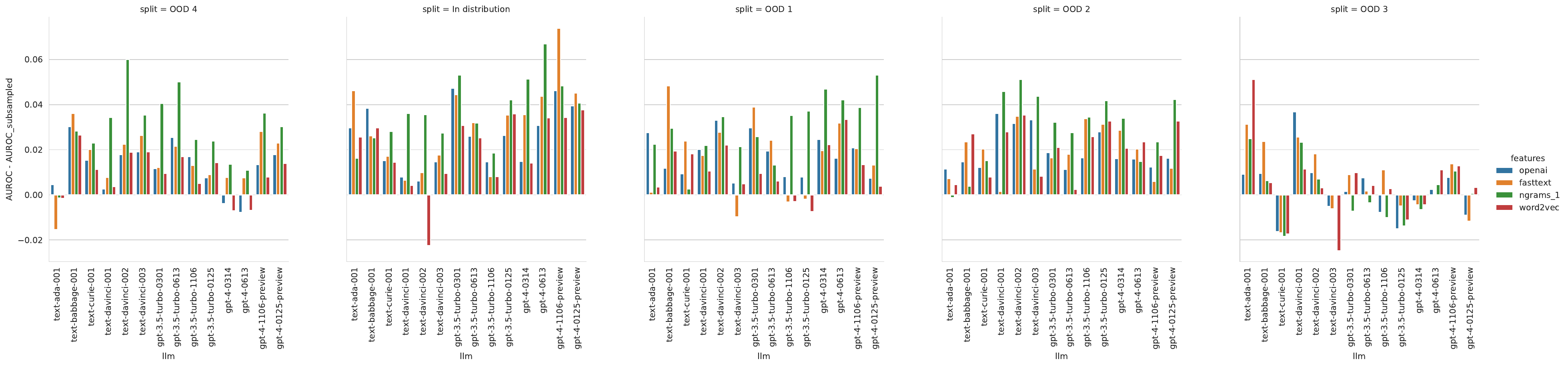}
        \caption{$\Dval$.}        
    \end{subfigure}
    \begin{subfigure}[b]{\textwidth}
        \includegraphics[width=\textwidth]{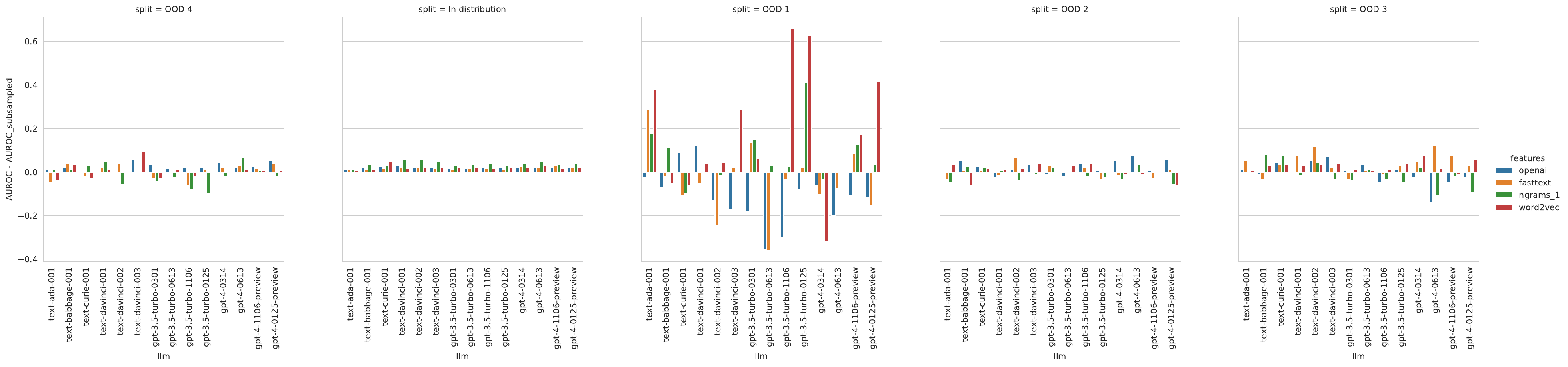}
        \caption{$\Dtest$.}        
    \end{subfigure}
    \caption{Difference between the AUC of a specific assessor trained on the full $\Dtrain$ and one trained on a random subsample of $\Dtrain$ of size 3000, for different choices of the random split for the KindsOfReasoning collection. Positive values indicate better performance of the specific assessor trained on the full $\Dtrain$, and viceversa. For each split and feature, various classifiers were trained on $\Dtrain$ and the best according to its performance on $\Dval$ was selected; the panels report the difference in performance of the latter on $\Dval$ and $\Dtest$.}
    \label{fig:other_figures_reasoning_difference}
\end{figure*}

Figure~\ref{fig:other_figures_reasoning_difference} shows the difference between the AUC of a specific assessor trained on the full $\Dtrain$ and one trained on a random subsample of $\Dtrain$ of size 3000, for different choices of the random split for the KindsOfReasoning collection. The difference is small on $\Dval$ (notice the $y$ scale of the graphs) and generally small for $\Dtest$ for all data splits, except for OOD 1, which reaches higher absolute values on both sides of 0.

\subsection{Impact of the number of reference points}
\label{app:n_ref}

\begin{figure*}
    \centering
    \begin{subfigure}[b]{\textwidth}
        \includegraphics[width=\textwidth]{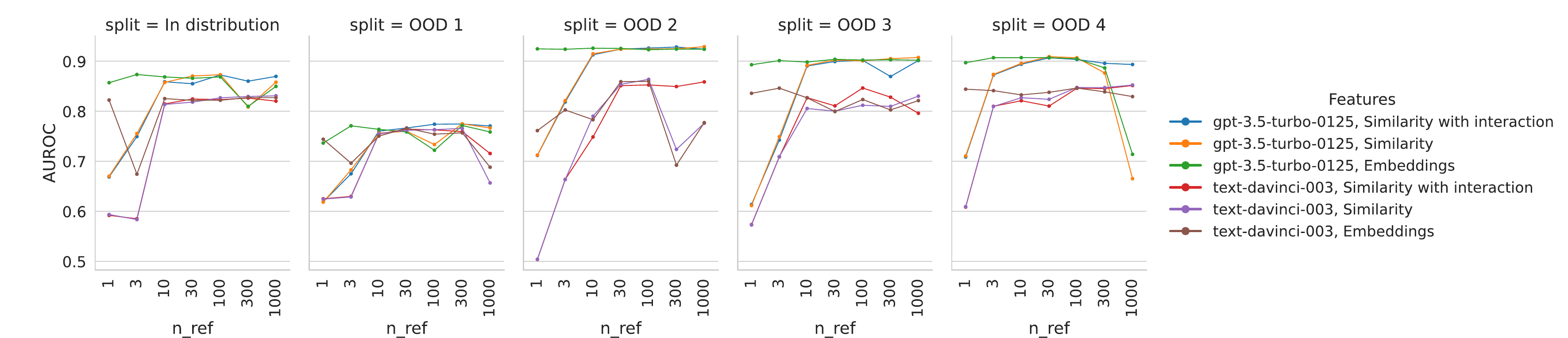}
        \caption{AUC with increasing number of reference instances on $\Dval$.}        
    \end{subfigure}
    \begin{subfigure}[b]{\textwidth}
        \includegraphics[width=\textwidth]{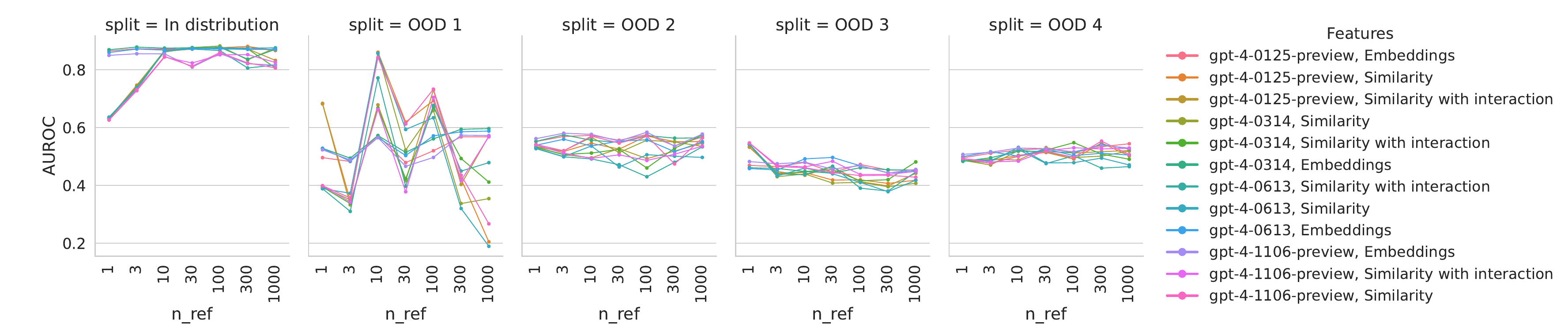}
        \caption{AUC with increasing number of reference instances on $\Dtest$.}        
    \end{subfigure}
    \caption{AUC for generic assessors trained with an increasing number reference instances in $\Dref$, for different splits on KindsOfReasoning. To produce this plot, only one possible selector (clustering on the OpenAI embeddings) and one classifier (XGBoost) were considered. For each split and number of reference instances, the results show the validation and test AUC with different choices of the instance-intrinsic features, for all test LLMs.}
    \label{fig:n_ref_reasoning}
\end{figure*}

\begin{figure*}
    \centering
    \begin{subfigure}[b]{\textwidth}
        \includegraphics[width=\textwidth]{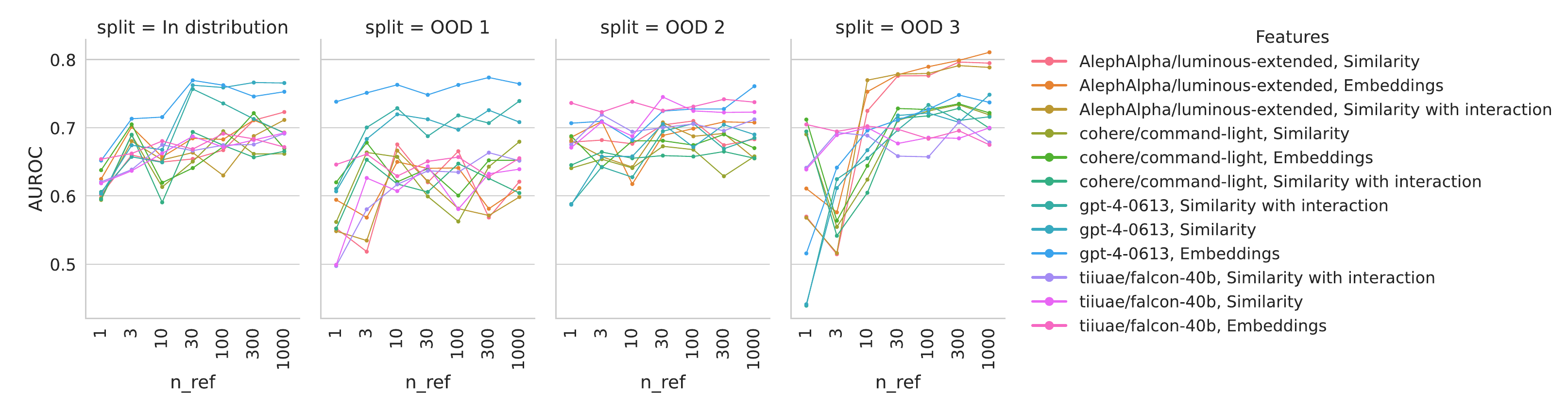}
        \caption{AUC with increasing number of reference instances on $\Dval$.}        
    \end{subfigure}
    \begin{subfigure}[b]{\textwidth}
        \includegraphics[width=\textwidth]{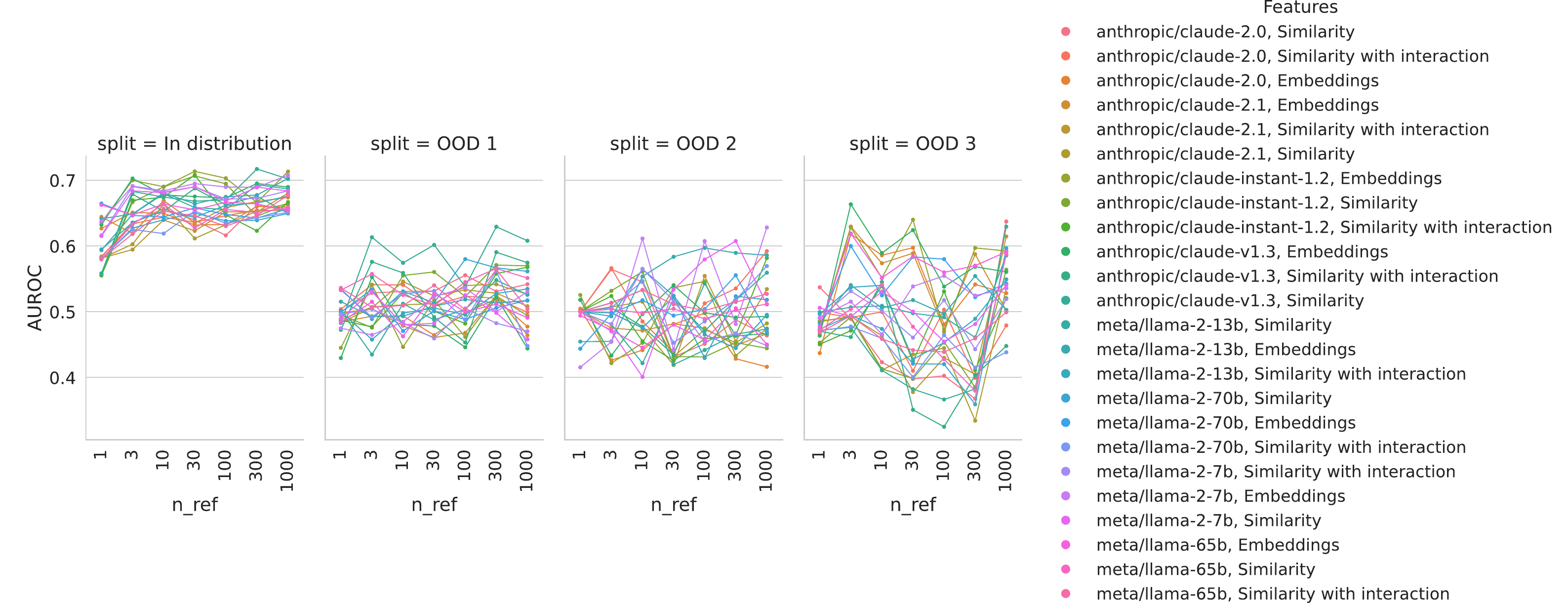}
        \caption{AUC with increasing number of reference instances on $\Dtest$.}        
    \end{subfigure}
    \caption{AUC for generic assessors trained with an increasing number reference instances in $\Dref$, for different splits on HELM-Lite. To produce this plot, only one possible selector (clustering on the OpenAI embeddings) and one classifier (XGBoost) were considered. For each split and number of reference instances, the results show the validation and test AUC with different choices of the instance-intrinsic features, for all test LLMs.}
    \label{fig:n_ref_helm}
\end{figure*}

Figures~\ref{fig:n_ref_reasoning} and \ref{fig:n_ref_helm} show the performance of the generic assessor in predicting the performance of $\Lval$ on $\Dval$ (top panels) and $\Ltest$ on $\Dtest$, for different values of the number of reference points selected, for different splits of the KindsOfReasoning and HELM-Lite collection respectively. In particular, this experiment was conducted by considering only one selector method (clustering on embeddings) and one base classifier (XGBoost); the results show the performances using various choices of the instance-intrinsic features. Broadly, it can be seen as the performance on $\Dval$ roughly peaks around 30 reference instances (although a few cases are roughly constant and some others show a drop for very high number of reference instances). No clear trend can instead be seen for the performance on $\Dtest$ for the OOD splits.

\end{document}